\pdfoutput=1
\documentclass[11pt]{article}

\usepackage{colortbl}
\usepackage{booktabs}
\usepackage[]{acl}

\usepackage{times}
\usepackage{latexsym}

\usepackage{graphicx}
\usepackage{booktabs}
\usepackage{multirow}

\usepackage[T1]{fontenc}
\usepackage[utf8]{inputenc}

\usepackage{microtype}

\usepackage{inconsolata}

\newcommand*\samethanks[1][\value{footnote}]{\footnotemark[#1]}

\title{On the Relationship between Sentence Analogy Identification and Sentence Structure Encoding in Large Language Models}

\author{
    Thilini Wijesiriwardene\textsuperscript{1}\thanks{~~Corresponding author}~,   
    Ruwan Wickramarachchi\textsuperscript{1},
    \bf{Aishwarya Naresh Reganti\textsuperscript{2}}\\
    \bf{Vinija Jain\textsuperscript{3,4}\thanks{~~Work does not relate to position at Amazon.} \,,
    Aman Chadha\textsuperscript{3,4}\samethanks \,\,, 
    Amit Sheth\textsuperscript{1},
    Amitava Das\textsuperscript{1}}\\ 
    \textsuperscript{1}AI Institute, University of South Carolina, USA,\\
    \textsuperscript{2}Carnegie Mellon University, Pittsburgh, USA, \\
    \textsuperscript{3}Amazon GenAI, USA, 
    \textsuperscript{4}Stanford University, USA\\
    \texttt{thilini@sc.edu}
    }

\begin{document}
\maketitle
\begin{abstract}
The ability of Large Language Models (LLMs) to encode syntactic and semantic structures of language is well examined in NLP. Additionally, analogy identification, in the form of word analogies are extensively studied in the last decade of language modeling literature. In this work we specifically look at how LLMs' abilities to capture sentence analogies (sentences that convey analogous meaning to each other) vary with LLMs' abilities to encode syntactic and semantic structures of sentences. Through our analysis, we find that LLMs' ability to identify sentence analogies is positively correlated with their ability to encode syntactic and semantic structures of sentences. Specifically, we find that the LLMs which capture syntactic structures better, also have higher abilities in identifying sentence analogies. 
\end{abstract}

\section{Introduction}

Analogies facilitate the transfer of meaning and knowledge from one domain to another. Making and identifying analogies is a central tenet in human cognition \cite{hofstadter2001analogy, holyoakplace} and is aided by humans' ability to process the structure of language. In the domain of NLP, several types of textual analogies are identified, such as word analogies \cite{yuan2023analogykb, gladkova-etal-2016-analogy, gao2014wordrep}, proportional word analogies \cite{chen2022kar, ushio-etal-2021-bert, szymanski2017temporal, drozd2016word}, sentence-analogies \cite{afantenos2021analogies, zhu-de-melo-2020-sentence, wang2020vector} and more recently analogies of procedural/long text \cite{sultan-shahaf-2022-life}. This work explicitly looks at sentence-level analogies which are sentence pairs that are analogues in meaning to each other \footnote{For more details on sentence analogies please refer to \cite{wijesiriwardene-etal-2023-analogical}}.

\begin{figure}[!ht]
\center
\includegraphics[width=.50\textwidth]{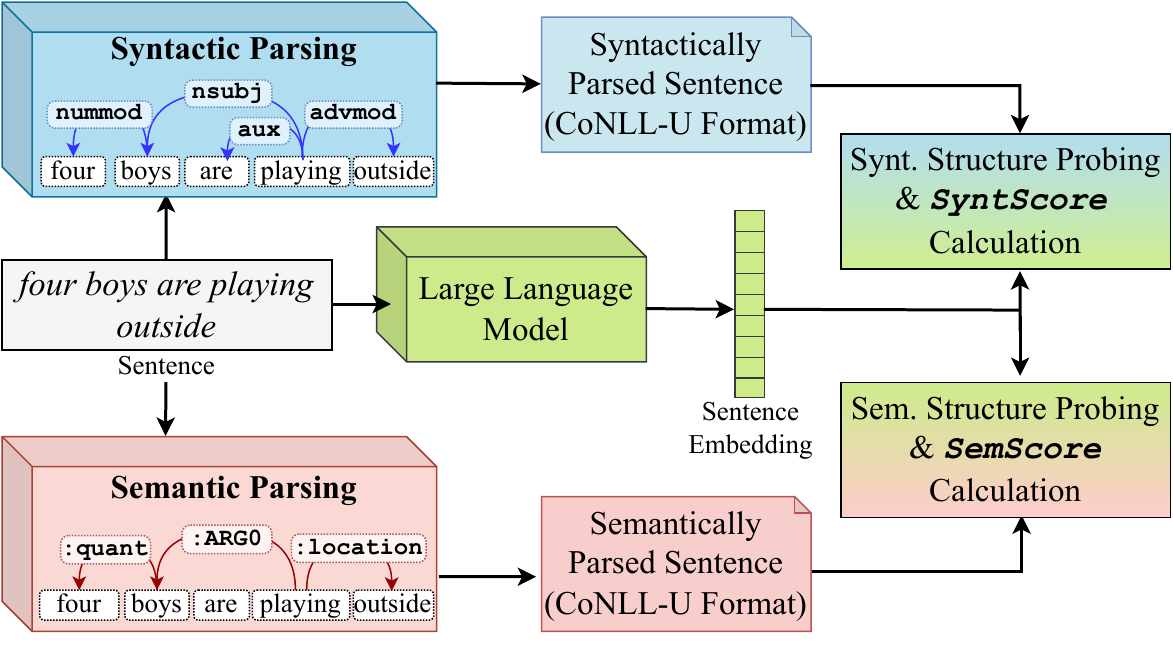}
\caption{This pipeline details the process of quantifying the LLMs abilities to capture sentence structure via \texttt{SyntScore} and \texttt{SemScore} values for a given sentence. In this work, we apply this process to a dataset of 100K sentences. The dataset is divided into 0.8 for training the structure probe and 0.1 for testing.}
\vspace{-1em}
\label{fig:pipeline}
\end{figure}

Despite the existence of several established benchmarks (e.g., SuperGLUE \cite{wang2019superglue} and GLUE \cite{wang-etal-2018-glue}) which evaluate the abilities of LLMs extrinsically, \citet{wijesiriwardene-etal-2023-analogical} propose a more challenging intrinsic benchmark that focuses on LLMs' ability to identify analogies across a range of complexities. 

Identification of analogies relies on the utilization of implicit relational knowledge embedded within the relational structure of language \cite{gentner1983structure}. 

In this work we aim to explore the relationship between sentence analogy identification abilities and syntactic and semantic structure encoding abilities of LLMs\footnote{Our code is available at: \url{https://github.com/Thiliniiw/llms-synt-struct-sentence-analogies}}.
 
Specifically, our main contribution is an analysis of the relationship between the analogy identification ability and sentence structure encoding abilities of LLMs. Additionally, we extend the sentence structure probing techniques introduced by \citet{hewitt-manning-2019-structural} (which only supports BERT and ELMo) to further work with encoder-decoder-based LLMs and LLMs that use two transformer architectures. Finally, we extend the structure probing technique originally used for syntactic structure probing in the novel context of semantic structure probing.

\section{Related Work}

Assessing the ability of Neural Networks (NN) to encode syntactic and semantic structures of language is well examined in NLP \cite{nivre2007conll, manning1999foundations, parsing2009speech}. \citet{everaert2015structures} emphasize that the meaning of sentences is inferred by the hierarchical structures provided by syntactic and semantic properties of language. 

%syntactic parsing 
Syntactic parsing aims to derive the syntactic dependencies in a sentence, such as subjects, objects, quantifiers, determiners and other similar elements. Early probing tasks \cite{adi2016fine, shi2016does} tried to identify NNs' abilities to capture syntactic structures by classifying sentences with single and plural subjects. Later, \citet{conneau2018you} showed that NNs could capture the maximal parse tree depth. The structure probing technique used and extended in this work \cite{hewitt-manning-2019-structural} is related but distinct due to its ability to implicitly capture the parse tree structures through simple distance measures between the vector representations of the words. 

%semantic parsing 
Compared to syntactic parsing, the NLP communities' interest in semantic parsing is growing. Semantic parsing maps natural language sentences to a complete, formal meaning representation. Semantic parsing is achieved via combining the Semantic Role Labelling (SRL) approaches with syntactic dependency parsing  \cite{hajic2009conll, surdeanu2008conll} and more recently via semantic dependency parsing \cite{oepen-etal-2014-semeval, oepen2015semeval}. This work uses the semantic dependency parsing approach based on mean field variational inference (MFVI) augmented with character and lemma level embeddings introduced by \citet{wang2019second}. 

\section{Approach}

Our approach to exploring the relationship between analogy identification and sentence structure encoding in LLMs is detailed in the following three subsections. We explain the dataset used, in Section \ref{sec:3.1}, the analogy identification abilities of LLMs in Section \ref{sec:3.2} and the sentence structure encoding abilities of LLMs in Section \ref{sec:3.3}.

\subsection{Dataset}
\label{sec:3.1}
We experiment on a dataset of 100K English sentences. Specifically, the dataset used in this work is randomly picked from the sentence corpus of levels three, four and five of the analogy taxonomy introduced in \cite{wijesiriwardene-etal-2023-analogical}. The composition of the dataset is presented in Table \ref{table:dataset_stats} (duplicates removed). Specifically, we obtain sentence-analogy pairs provided by \citet{wijesiriwardene-etal-2023-analogical} and split the pairs to obtain single sentences used in this work.

\begin{table}[]
\resizebox{\columnwidth}{!}{%
\begin{tabular}{@{}lcr@{}}
\toprule
\textbf{Analogy Taxo. Level}  & \textbf{Datasets} & \textbf{\# Sentences}         \\ \midrule
Level Three                 &  Random deletion/masking/reorder        & 69,111                \\
Level Four                  &  Negation        & 1,245                 \\
Level Five                  &  Entailment       & 29,644                \\ \midrule
\multicolumn{2}{l}{\textbf{Total \# Sentences}} & \textbf{100,000} \\ \bottomrule
\end{tabular}%
}
\vspace{-0.5em}
\caption{Dataset statistics.}
\vspace{-1.2em}
\label{table:dataset_stats}
\end{table}

\subsection{Large Language Models and their Ability to Capture Sentence Analogies}
\label{sec:3.2}
We experiment on the eight language models used in a study by \citet{wijesiriwardene-etal-2023-analogical} namely, BERT \cite{devlin2018bert}, RoBERTa \cite{liu2019roberta}, ALBERT \cite{lan2019albert}, LinkBERT \cite{yasunaga2022linkbert}, SpanBERT \cite{joshi2020spanbert} and XLNet \cite{yang2019xlnet} which are encoder-based LLMs, T5 \cite{raffel2020exploring}, an encoder-decoder-based LLM and ELECTRA \cite{clark2020electra}, an LLM based on two transformer architectures. We refer readers to cited publications for details on the specific LLMs.

\citet{wijesiriwardene-etal-2023-analogical} introduced a taxonomy of analogies starting from less complex word-level analogies to more complex paragraph-level analogies and evaluated how each LLM performs on identifying analogies at each level of the taxonomy. An analogy is a pair of lexical items that are identified to hold a similar meaning to each other. Therefore the distance between a pair of analogous lexical items in the vector space should be smaller. The same authors identify Mahalanobis Distance (MD) \cite{mahalanobis1936generalized} to be a better measurement of the distance between two analogous sentences in the vector space. Therefore in this work, the ability of each LLM to identify sentence analogies is represented by the mean MD calculated for the sentence-level datasets (levels 3, 4 and 5) present in the analogy taxonomy. These mean values are calculated based on the reported values by \citet{wijesiriwardene-etal-2023-analogical}.

\subsection{Large Language Models and their Ability to Capture Sentence Structures}
\label{sec:3.3}
\citet{hewitt-manning-2019-structural} introduced a probing approach to evaluate whether syntax trees (sentence structures) are encoded in Language Models' (LMs') vector geometry. The probing model is trained on  train/dev/test splits of the Penn Treebank \cite{marcus-etal-1993-building} and tested on both BERT \cite{devlin2018bert} and ELMo \cite{peters-etal-2018-deep}. An LM's ability to capture sentence structure is quantified by its ability to correctly encode the gold parse tree (provided in the Penn Treebank dataset) within its embeddings for a given sentence.

The authors introduce a path distance metric and a path depth metric for evaluation. The distance metric captures the path length between each pair of words measured by Undirected Unlabeled Attachment Score (UUAS) and average Spearman correlation of true to predicted distances (DSpr). The depth metric evaluates the model's ability to identify a sentence's root, measured as root accuracy percentage. Additionally, the depth metric also evaluates the ability of the model to recreate the word order based on their depth in the parse tree identified as Norm Spearman (NSpr.)\footnote{We do not use NSpr. in this work.} We refer the readers to \citet{hewitt-manning-2019-structural} for further details on the technique and evaluation metrics.

\section{Experimental Setup}

\textls[-10]{Exploring the relationship between analogy identification and sentence structure encoding abilities of LLMs requires a representative score to quantify (i) analogy identification ability (\texttt{AnalogyScore)}, (ii) semantic structure identification ability \texttt{(SemScore)}, and (iii) syntactic structure identification ability \texttt{(SyntScore)} of each LLM.}

We obtain \texttt{AnalogyScore} by calculating the means of reported MD measures obtained for each sentence-level dataset in \citet{wijesiriwardene-etal-2023-analogical}. 

To obtain the \texttt{SemScore} (see Figure \ref{fig:pipeline}), we first parse all the sentences in our dataset using the MFVI approach \cite{wang2019second}. The resulting semantically parsed sentences (in CoNLL-U format)\footnote{\url{https://universaldependencies.org/format.html}} and the  LLM embeddings of the original sentences are then sent to the structure probing technique \cite{hewitt-manning-2019-structural}. The structure probe is trained on 80K sentences from the dataset and the DSpr and UUAS values representing parse distance and root accuracy (RootAcc) value representing parse depth are reported on the test split with 10K sentences. Finally, the \texttt{SemScore} is computed as a combined score by taking the mean of the z-score normalizations of these three measures \(Z_{DSpr}, Z_{UUAS}, Z_{RootAcc}\) (see Table \ref{tab:zscores}).

\[\texttt{SemScore} = \frac{1}{3} (Z_{DSpr} + Z_{UUAS} + Z_{RootAcc})\]

To obtain the \texttt{SyntScore} (see Figure \ref{fig:pipeline}), we follow the same steps but parse the sentences syntactically.
%(in CoNLL-U format). 
Finally, we calculate the Spearman's rank correlation (SRC) and Kendall's rank correlation (KRC) between \texttt{AnalogyScore} and \texttt{SyntScore}, as well as \texttt{AnalogyScore} and \texttt{SemScore}. %Implementation details are discussed in appendix \ref{sec:implementation details}.

\subsection{Implementation Details}
When extending the structure probing technique by \citet{hewitt-manning-2019-structural} to facilitate additional LLMs, we use the HuggingFace implementation\footnote{\url{https://huggingface.co/models}} of the LLMs. For semantic parsing, we use the trained mean field variational inference (MFVI) model augmented with character and lemma-level embeddings provided by the SuPar\footnote{\url{https://github.com/yzhangcs/parser}}. For syntactic parsing of the sentences we employ Stanford CoNLL-U dependency parser\footnote{\url{https://nlp.stanford.edu/software/nndep.html}}.

\begin{table*}[]
\resizebox{\textwidth}{!}{%
\begin{tabular}{@{}lcccccccccccc@{}}
\toprule
\multirow{4}{*}{\textbf{Model}} & \multicolumn{6}{c}{\textbf{Original Scores}}                & \multicolumn{6}{c}{\textbf{Normalized Scores}}         \\ \cmidrule(l){2-13} 
 &
  \multicolumn{3}{c}{\textbf{Syntactic}} &
  \multicolumn{3}{c}{\textbf{Semantic}} &
  \multicolumn{3}{c}{\textbf{Syntactic}} &
  \multicolumn{3}{c}{\textbf{Semantic}} \\
 &
  \multicolumn{2}{c}{\textbf{Distance}} &
  \textbf{Depth} &
  \multicolumn{2}{c}{\textbf{Distance}} &
  \textbf{Depth} &
  \multicolumn{2}{c}{\textbf{Distance}} &
  \textbf{Depth} &
  \multicolumn{2}{c}{\textbf{Distance}} &
  \textbf{Depth} \\
                       & DSpr & UUAS & RootAcc & DSpr & UUAS & RootAcc  & $Z_{DSpr}$  & $Z_{UUAS}$  & $Z_{RootAccu}$ &  $Z_{DSpr}$  & $Z_{UUAS}$  & $Z_{RootAccu}$ \\\midrule
ALBERT                 & 0.59 & 0.46 & 0.35       & 0.38 & 0.13 & 0.19       & -1.56 & -2.30 & -2.58 & 0.39  & -1.30 & 0.36  \\
BERT                   & 0.73 & 0.72 & 0.74       & 0.38 & 0.16 & 0.17       & 0.87  & 0.62  & 0.56  & 0.39  & -0.03 & 0.07  \\
Electra                & 0.70 & 0.76 & 0.75       & 0.38 & 0.14 & 0.15       & 0.34  & 1.01  & 0.63  & 0.39  & -0.73 & -0.28 \\
LinkBERT               & 0.70 & 0.68 & 0.69       & 0.38 & 0.15 & 0.05       & 0.33  & 0.18  & 0.15  & 0.37  & -0.27 & -1.79 \\
RoBERTa                & 0.74 & 0.74 & 0.73       & 0.38 & 0.16 & 0.29       & 1.06  & 0.77  & 0.49  & 0.37  & 0.25  & 1.89  \\
SpanBERT               & 0.74 & 0.72 & 0.74       & 0.38 & 0.14 & 0.20       & 1.06  & 0.56  & 0.55  & 0.37  & -0.97 & 0.54  \\
T5                     & 0.63 & 0.64 & 0.71       & 0.37 & 0.19 & 0.17       & -0.79 & -0.31 & 0.28  & -2.65 & 1.64  & 0.05  \\
XLNet                  & 0.60 & 0.62 & 0.66       & 0.38 & 0.18 & 0.11       & -1.31 & -0.53 & -0.08 & 0.37  & 1.42  & -0.83 \\ \bottomrule
\end{tabular}%
}
\caption{DSpr, UUAS measures indicating Parse Distance (Distance) and RootAcc measure indicating Parse Depth (Depth). Original Scores denote original output values of the structure probe technique and Normalized Scores are z-score normalized. Higher values indicate a stronger ability of the LLMs to capture sentence structures.}
\vspace{-1em}
\label{tab:zscores}
\end{table*}

% Table without COLORS --- FINAL
\begin{table}[]
\centering
\resizebox{\columnwidth}{!}{%
\begin{tabular}{@{}lcccccc@{}}
\toprule
\multirow{2}{*}{\textbf{Model}} & \multicolumn{2}{c}{\textbf{AnalogyScore}} & \multicolumn{2}{c}{\textbf{SyntScore}} & \multicolumn{2}{c}{\textbf{SemScore}} \\
         & Score & Rank & Score & Rank & Score & Rank \\ \midrule
ALBERT   & 0.645  & 7    & -2.14 & 8    & -0.19 & 5    \\
BERT     & 0.505  & 3    & 0.68  & 3    & 0.14  & 3    \\
Electra  & 0.516  & 4    & 0.66  & 4    & -0.21 & 6    \\
LinkBERT & 0.608  & 6    & 0.22  & 5    & -0.56 & 8    \\
\textbf{RoBERTa}  & \textbf{0.458}  & \textbf{1}    & \textbf{0.78 } & \textbf{1}    & \textbf{0.84}  & \textbf{1}    \\
SpanBERT & 0.461  & 2    & 0.72  & 2    & -0.02 & 4    \\
T5       & 0.524  & 5    & -0.27 & 6    & -0.32 & 7    \\
XLNet    & 0.747  & 8    & -0.64 & 7    & 0.32  & 2    \\ \bottomrule
\end{tabular}%
}
\caption{The values for \texttt{AnalogyScore}, \texttt{SyntScore} and \texttt{SemScore} and their corresponding rank values. \texttt{AnalogyScore} ranges between [0,1], 0 being the best. For \texttt{SyntScore} and \texttt{SemScore} higher the values better the ability of LLMs to capture sentence structure.}
\vspace{-1em}
\label{tab:analogyscore}
\end{table}

% Table with COLORS
\begin{table}[]
\centering
\resizebox{\columnwidth}{!}{%
\begin{tabular}{@{}lcccccc@{}}
\toprule
\multirow{2}{*}{\textbf{Model}} & \multicolumn{2}{c}{\textbf{AnalogyScore}} & \multicolumn{2}{c}{\textbf{SyntScore}} & \multicolumn{2}{c}{\textbf{SemScore}} \\

         & Score & Rank & Score & Rank & Score & Rank \\ \midrule
AlBERT   & 0.645  & 7  & -2.14 & 8     & -0.19  & 5  \\
BERT     & 0.505 & 3  & 0.68  & 3 & 0.14  & 3   \\
Electra  & 0.516 & 4  & 0.66  & 4 & -0.21  & 6  \\
LinkBERT & 0.608 & 6 & 0.22  & 5 & -0.56 & 8  \\
\textbf{RoBERTa}  & \textbf{0.458 }& \textbf{1 }& \textbf{0.78 } & \textbf{1}  & \textbf{0.84}  & \textbf{1 } \\
SpanBERT & 0.461 & 2  & 0.72  & 2 & -0.02 & 4   \\
T5       & 0.524 & 5 & -0.27  & 6 & -0.32  & 7   \\
XLNet    & 0.747 & 8 & -0.64  & 7 & 0.32  & 2   \\
\bottomrule
\end{tabular}
}
\caption{The values for \texttt{AnalogyScore}, \texttt{SyntScore} and \texttt{SemScore} and their corresponding rank values. \texttt{AnalogyScore} ranges between [0,1], 0 being the best. For \texttt{SyntScore} and \texttt{SemScore} higher the values better the ability of LLMs to capture sentence structure.}
\end{table}

\section{Results}
In this section, we look at the findings of this work with regard to semantic and syntactic structure encoding abilities and analogy identification abilities of LLMs.

\subsection{Semantic and Syntactic Structure Encoding Abilities of LLMs}
  
We tabulate the structure probing results in original metrics (Table \ref{tab:zscores}) and the performance of each LLM in identifying sentence analogies and capturing the semantic and syntactic structures (Table \ref{tab:analogyscore}). It is interesting to note that RoBERTa,  the best-performing LLM for analogy identification (\texttt{AnalogyScore} $ = 0.458$), holds the highest \texttt{SyntScore} and \texttt{SemScore}. XLNet is the lowest-performing model for analogy identification as well as syntactic structure identification. Yet it performs second-best in semantic structure identification. SpanBERT ranks second in both analogy identification and syntactic structure identification but holds the median \texttt{SemScore}.   

\subsection{Analogy Identification and Syntactic Structure Encoding Abilities of LLMs}

\textls[-5]{We use SRC and KRC values to analyze the correlation between LLMs' ability to identify sentence analogies denoted by \texttt{AnalogyScore} and LLMs' ability to encode syntactic structures of sentences denoted by \texttt{SyntScore}. Both correlation measures show a significant positive correlation between \texttt{AnalogyScore} and \texttt{SyntScore}. Specifically, the SRC between \texttt{AnalogyScore} and \texttt{SyntScore} is 0.95 ($p < 0.001$). The KRC between \texttt{AnalogyScore} and \texttt{SyntScore} is 0.86 ($p = 0.002$).}

\subsection{Analogy Identification and Semantic Structure Encoding abilities of LLMs}

\textls[-2]{Similar to the previous section, we compute the SRC and KRC values to asses the correlations between \texttt{AnalogyScore} and \texttt{SemScore}. We see that both correlations are positive with SRC of 0.33 ($p=0.42$) and KRC of 0.28 ($p=0.40$) between \texttt{AnalogyScore} and \texttt{SemScore}}. %We consider these particular p values to be significant given the low number of observations (8 LLMs).}

\section{Limitations}

Several contemporary probing techniques, such as those outlined in \citet{voita-titov-2020-information} and \citet{pimentel-etal-2020-pareto}, have emerged subsequent to the methodology employed in the present investigation \cite{hewitt-manning-2019-structural}. Nevertheless, in the context of our current study, we have only chosen to employ \cite{hewitt-manning-2019-structural} owing to its adaptable nature, which facilitates extension to various LLMs that are of particular interest to our current research.

Even though Abstract Meaning Representation (AMR) \cite{banarescu2013abstract} is a popular and widely used technique to parse sentences semantically, in current work, we use MFVI, a semantic parsing approach introduced by \citet{wang2019second} because of the limitations posed by the structure probing technique used \cite{hewitt-manning-2019-structural}. This technique requires the mapped LLM embeddings and semantic dependency parsed sentences to be of the same length. However, as it is known, AMRs abstract away from the syntactic idiosyncrasies of the language and overlook certain auxiliary words from the parse results, limiting its use in this work.

 The present study employs a semantic parsing technique reported to exhibit a high accuracy level of 94\% \cite{wang2019second}. However, it is important to note that for the purposes of our investigation, we make the assumption that the semantically parsed sentences generated by this particular method are entirely accurate, thereby employing them as the gold standard data. It is worth mentioning that this choice may introduce some degree of bias into our examination of the semantic structure probing.

\section{Conclusion and Future Directions}
This work explores the relationship between LLMs' ability to identify sentence analogies and encode sentence structures in their embeddings. Through detailed experiments, we show that the sentence analogy identification ability of LLMs is positively correlated with their ability to encode syntactic and semantic structures of sentences. Particularly, LLMs that better capture syntactic structures have a higher correlation to analogy identification. In summary this work explores how LLMS utilize the knowledge of semantic and syntactic structures of sentences to identify analogies. Moving forward, we aim to explore the potential of extending the current approach to enhance explainability of LLMs within the broader domain of NLP.

\section*{Acknowledgements}
We thank the anonymous reviewers for their constructive comments. This work was supported in part by the NSF grant \#2335967: EAGER: Knowledge-guided neurosymbolic AI with guardrails for safe virtual health assistants. Any opinions, findings, conclusions, or recommendations expressed in this material are those of the authors and do not necessarily reflect the views of the funding organization.

\bibliography{anthology,custom}
\end{document}